%% file: main.tex
\documentclass[10pt, conference]{IEEEtran}

\IEEEoverridecommandlockouts
\usepackage{cite}
\usepackage{graphicx}
\usepackage{textcomp}
\usepackage{xcolor}
\usepackage[a4paper, total={184mm,239mm}]{geometry}

\input{packages}
\input{commands}

\def\BibTeX{{\rm B\kern-.05em{\sc i\kern-.025em b}\kern-.08em
    T\kern-.1667em\lower.7ex\hbox{E}\kern-.125emX}}

\begin{document}

\title{Pipe-BD: Pipelined Parallel Blockwise Distillation}



\author{%
\IEEEauthorblockN{Hongsun Jang\IEEEauthorrefmark{2}, Jaewon Jung\IEEEauthorrefmark{4}, Jaeyong Song\IEEEauthorrefmark{3}, Joonsang Yu\IEEEauthorrefmark{5}, Youngsok Kim\IEEEauthorrefmark{3}, and Jinho Lee\IEEEauthorrefmark{2}$^*$}\vspace{2pt}
\IEEEauthorblockA{\IEEEauthorrefmark{2}Department of Electrical and Computer Engineering, Seoul National University}
\IEEEauthorblockA{\IEEEauthorrefmark{4}Department of Artificial Intelligence, Yonsei University}
\IEEEauthorblockA{\IEEEauthorrefmark{3}Department of Computer Science, Yonsei University}
\IEEEauthorblockA{\IEEEauthorrefmark{5}CLOVA ImageVision, CLOVA AI Lab, NAVER}
\vspace{2pt}
\IEEEauthorblockA{hongsun.jang@snu.ac.kr, \{jungjaewon, jaeyong.song\}@yonsei.ac.kr, joonsang.yu@navercorp.com,} 
\IEEEauthorblockA{youngsok@yonsei.ac.kr, leejinho@snu.ac.kr}

\thanks{* Corresponding author}
}


\maketitle

\begin{abstract}
Training large deep neural network models is highly challenging due to their tremendous computational and memory requirements.
Blockwise distillation provides one promising method towards faster convergence by splitting a large model into multiple smaller models.
In state-of-the-art blockwise distillation methods, training is performed block-by-block in a data-parallel manner using multiple GPUs. 
To produce inputs for the student blocks, the teacher model is executed from the beginning until the current block under training.
However, this results in a high overhead of redundant teacher execution, low GPU utilization, and extra data loading.
To address these problems, we propose \scheme, a novel parallelization method for blockwise distillation. 
\scheme aggressively utilizes pipeline parallelism for blockwise distillation, eliminating redundant teacher block execution and increasing per-device batch size for better resource utilization.
We also extend to hybrid parallelism for efficient workload balancing.
As a result, \scheme achieves significant acceleration without modifying the mathematical formulation of blockwise distillation. 
We implement \scheme on PyTorch, and experiments reveal that \scheme is effective on multiple scenarios, models, and datasets.

\end{abstract}

\begin{IEEEkeywords}
Distributed Training, Knowledge Distillation, Neural Architecture Search, Model Compression
\end{IEEEkeywords}

\section{Introduction}

Modern deep neural network models are known to incur huge computational and memory requirements, especially with large-scale datasets~\cite{imagenet}.
With the continuing growth in model size, it takes tens, if not hundreds, of GPU days to train them~\cite{imagenethour}, and the model size often exceeds the GPU memory capacity. 
Especially for methods that explore large solution spaces such as the neural architecture search (NAS)~\cite{proxyless, liu2019darts}, the problem becomes even more significant.
This problem mandates the use of model parallelism~\cite{gpipe, naspipe}, which creates substantial throughput loss with inevitable pipeline bubbles.

\begin{figure}[t]
\centering
 \includegraphics[width=0.8\columnwidth]{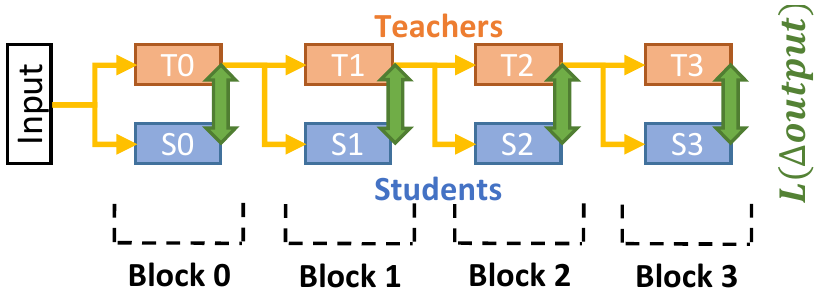}
 \caption{Conceptual diagram of blockwise distillation.
 }
 \label{fig:background}
\end{figure}

\Bd~\cite{pbd,progressive_bd,dna} is one promising approach to mitigate such problems.
As illustrated in \cref{fig:background}, \bd splits the model into multiple smaller blocks. 
As opposed to traditional knowledge distillation methods that rely on input data and output labels from both ends, \bd uses the intermediate activation values of pretrained blocks of a `teacher' to train each `student' block.
As a result, each block converges faster (i.e., fewer epochs) due to the smaller solution space.

Contrary to the earlier belief that teachers must be larger than students, recent studies have revealed that smaller teachers can be used to train larger students~\cite{distillation_regularization}.  
With such findings, \bd is used in various fields such as model compression~\cite{pbd, yu2019recasting} 
and NAS~\cite{dna, donna}.
Since training a small teacher for a new task is quick and easy, \bd can be applied in most cases where traditional training is used.

However, the existing state-of-the-art methods for \bd~\cite{dna,pbd} exhibit several inefficiencies.
Relying on the traditional data-parallel training scheme, 
they train each student block one by one independently. 
While this fully exploits the independent nature of the blocks, it is not the best choice for training throughput.
First, to train a single intermediate student block, the teacher blocks must be executed from the beginning to the designated block.
As a result, the teacher blocks exhibit substantial redundant execution, especially with blocks closer to the output.
Second, with data parallelism, a batch of data is split among multiple GPUs, which leads to a smaller batch size per GPU, often resulting in resource under-utilization.
Some approaches use a larger batch size to mitigate this~\cite{imagenethour}, but it is known to be difficult to ensure model convergence~\cite{largebatch}.
Last, the data must be redundantly loaded for each student block. 
Unless the entire dataset fits into the GPU memory, the data are loaded from the CPU memory or disks.
As the memory and disks are shared system-wide, the extra data loading becomes another significant overhead in training.

To address the issues, we propose \emph{\scheme}, a novel parallel training method for \bd. 
We assign individual student blocks to different devices and compute a teacher network in a relayed manner, which can reduce teacher redundancy. 
Inspired by approaches with pipeline parallelism~\cite{gpipe,pipedream,megatron-lm,optimus}, we restructure the training schedule of the student blocks such that the training time is greatly improved. 

\scheme comprises three components:
First, we propose \emph{\tr}. 
Instead of relying on data parallelism, we spread the student model to multiple training devices (i.e., GPUs) in a block granularity. 
Then, blocks of the teacher model are executed by relaying the intermediate activation values between the devices. 
This approach has the advantages of eliminating extra data loading and increasing resource utilization from larger batch size per device.
Second, we propose \emph{\sks} to remove the scheduling bubbles and enhance the overall utilization. 
With \tr, 
devices have to wait for the intermediate activation values from previous devices, creating scheduling bubbles.
\Sks performs model parameter updates in a misaligned manner and starts the next step right ahead, so those bubbles can be removed.
Third, we suggest \emph{\dr}. 
Achieving a balance between devices is difficult with \bd because
of the limited number of blocks available in typical neural network structures. 
\Dr enables fine-grained balancing with further splitting blocks along the batch size dimensions. 


\scheme is implemented on PyTorch and can automatically make all scheduling decisions to improve the throughput.
Our extensive set of experiments shows \scheme achieves a significant speedup over the state-of-the-art methods on multiple use cases and environments ranging from 2.37$\times$ to 7.38$\times$.

\section{Background and Related Work}
\subsection{Blockwise Distillation}
Blockwise distillation~\cite{dna, pbd, progressive_bd} is a promising direction for training a neural network.
In traditional knowledge distillation, a student model is trained against a pre-trained teacher model. 
Because the solution space size is identical to that of conventional supervised training, it faces convergence and training time problems.
\Bd splits the larger teacher model into smaller ones and trains them blockwise as depicted in~\cref{fig:background}.
Each teacher block ($T_i$) and student block ($S_i$) pair obtains activation values from the previous teacher block ($T_{i-1}$).
This pair performs forward pass using the activation as input and creates teacher output activation and student output activation.
\Bd minimizes a loss function ($L(\Delta output)$) which measures the difference between these two activations, to distill knowledge from a teacher block to the dedicated student block.
This \bd process makes target problem spaces smaller and is known to converge faster. Many applications such as NAS~\cite{dna,donna} and model compression~\cite{pbd,yu2019recasting} use \bd because of these characteristics.

\subsection{Parallelization Baseline of Blockwise Distillation}
\label{sec:related_parallel}
State-of-the-art methods of \bd~\cite{dna} use the traditional data-parallel scheme to further accelerate the training as illustrated in~\cref{fig:scheme:baseline}.
This scheme trains a student block ($S_i$) with all devices in a data-parallel manner for fixed $n$ epochs, then moves on to train the next student block ($S_{i+1}$).
It redundantly loads data multiple times  because of this iterative training.
Each student block ($S_i$) requires the activation values from the previous teacher ($T_{i-1}$), so it also entails redundant teacher executions. Furthermore, it uses a smaller batch size per device which leads to under-utilization. Due to these inefficiencies, the data-parallel \bd suffers from poor scalability. 
An alternative scheme~\cite{pbd} regards the training of each layer as a single task and adopts bin packing algorithm to balance the workload.
However, it still has redundant teacher executions and suffers from workload imbalance when there are insufficient layers in the model.

\section{Motivation}
\label{sec:moti}

\begin{figure}
\centering
 \includegraphics[width=\columnwidth]{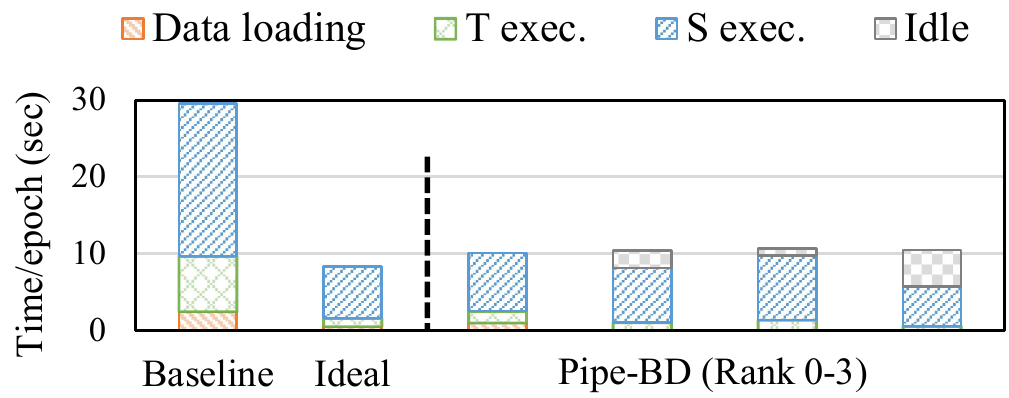}
 \caption{Motivational experiment. The breakdown demonstrates three major inefficiencies of baseline; redundant teacher execution, extra data loading, and low resource utilization.}
 \label{fig:motivation}
\end{figure}

In this section, we provide a motivational study highlighting the inefficiency of the existing parallel \bd training scheme and the need for a new approach.
%
%
\cref{fig:motivation} depicts the breakdown of time spent in parallel \bd with four RTX A6000 GPUs (NAS with Cifar-10; see \cref{sec:setup} for the detailed setup).
`Baseline' refers to the state-of-the-art parallel \bd method~\cite{dna}, where each block is trained sequentially using four devices with data parallelism.
As displayed in the chart, the training time is spent on data loading, teacher execution (forward pass), and student execution (forward/backward pass).

However, all three parts exhibit significant inefficiency, slowing down the training. To demonstrate the inefficiencies, we plot the `ideal' bar in \cref{fig:motivation} by measuring the training time of each part separately with a single GPU and dividing each time by four. 
This represents an imaginary system with perfect parallelization and infinite device memory.

\begin{figure}
\centering
\begin{subfigure}[t]{\columnwidth}
\includegraphics[width=\columnwidth]{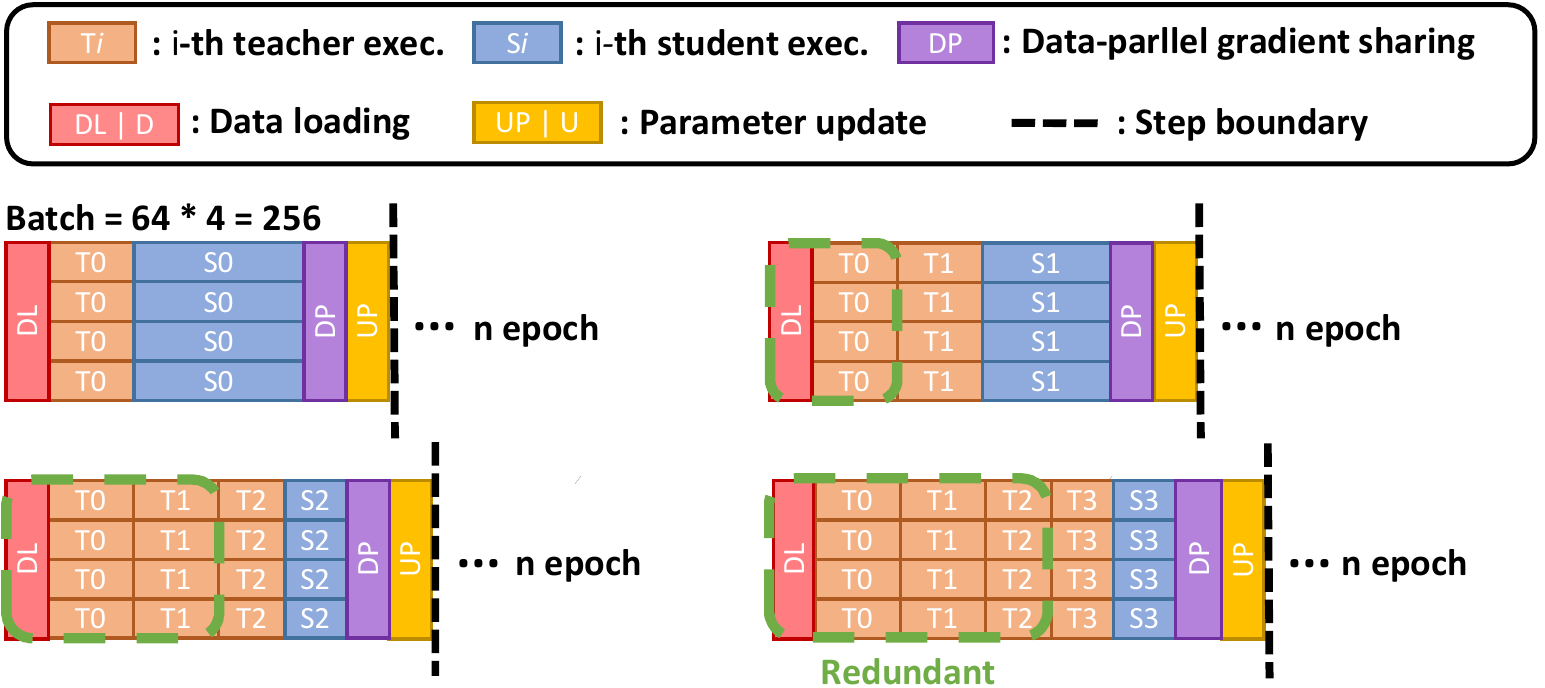}
\caption{Baseline}
\label{fig:scheme:baseline}
\end{subfigure}

\centering
\begin{subfigure}[t]{\columnwidth}
\includegraphics[width=\columnwidth]{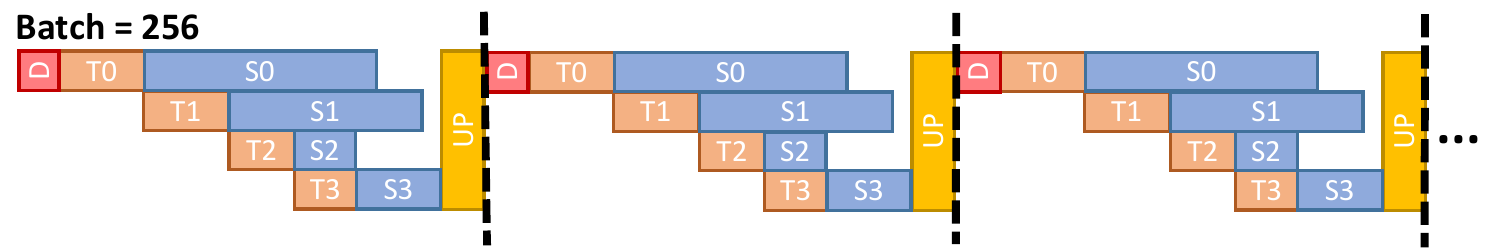}
\caption{w/ \TR}
\label{fig:scheme:ts}
\end{subfigure}
\centering

\begin{subfigure}[t]{\columnwidth}
\includegraphics[width=\columnwidth]{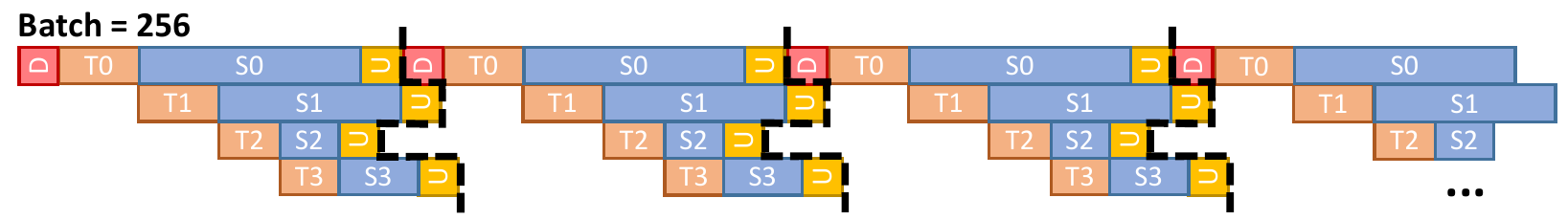}
\caption{w/ \SKS}
\label{fig:scheme:sks}
\end{subfigure}

\centering
\begin{subfigure}[t]{\columnwidth}
\includegraphics[width=\columnwidth]{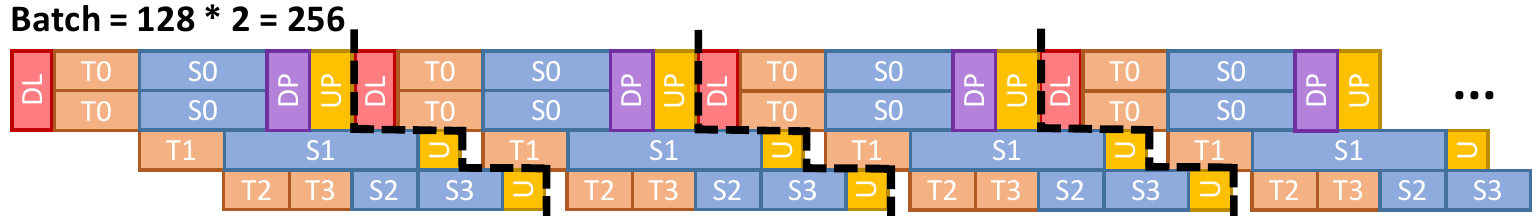}
\caption{w/ \DR}
\label{fig:scheme:dr}
\end{subfigure}

\caption{Illustration of the techniques in \scheme.}
 \label{fig:scheme}
\end{figure}

The large gaps in teacher execution and data loading time occur because the baseline has many redundant teacher executions and extra data loading.
Because each student block to train requires executing the teacher model from the beginning, the earlier teacher blocks are redundantly executed multiple times (see \cref{fig:scheme:baseline}).
Similarly, block-by-block training forces loading data as many as the number of blocks.
In addition, data-parallelism leads to smaller batch size per device, resulting in lower resource utilization.
As demonstrated in several empirical studies~\cite{GPUperformance,speedcomparison}, a sufficient per-device batch size is critical for training throughput, which is the cause of the gap on student execution time.
\scheme targets these inefficiencies.
As presented in \cref{fig:motivation}, \scheme reduces the training time close to the ideal case, with only a small overhead (idle).



\section{\scheme Method}

\subsection{\TR}
\scheme starts by restructuring the training pipeline of \bd with \tr.
As opposed to the baseline (\cref{fig:scheme:baseline}) where a single block is fully trained in a data-parallel manner before moving on to the next,
\tr exclusively distributes the teacher and student blocks to all training devices.
Then, each device relays the intermediate teacher activation values to the next device as depicted in \cref{fig:scheme:ts}.
The received activation is the input for both the teacher and the student block.
The teacher block is executed first, whose output activation is sent to the next device such that the execution of the next block can start.
Overlapped with the transmission, the forward pass execution of the student starts, taking the same input as the teacher block.
After calculating the loss by comparing the output activations of the teacher and the student, the backward pass of the student follows.
After all the backward passes are finished, parameter updates are performed on each block, completing the training step.

The \tr scheme has two advantages over the existing approach.
First, each device executes the stages with larger batches and enjoys better resource utilization.
For example, in the baseline using four devices with an effective batch size of 256, each device executes with a batch size of 64, which is often too small to fully utilize the hardware resources.
In contrast, with \tr, each device would run with a full batch size of 256, increasing resource utilization.
Second, the overhead of data loading is reduced. 
When the dataset is large, the data must come from the main memory or the disk, where both are system-wide shared resources.
Because \tr does not go through multiple training passes, the number of data loading decreases, leading to higher throughput.

One minor trade-off is communication overhead. 
In the baseline, gradient sharing must occur after every backward pass.
With \tr, there is some communication delay from relaying the intermediate activation values from one device to another.
However, the communication time is almost negligible in our target settings of single-node multi-device training. Furthermore, in both cases, most of the communications overlapped with computations.

\subsection{\SKS}
Although \tr removes the redundant teacher executions, the removed redundancy is not directly translated to speedup. 
At the beginning of each step, each device has to wait until the previous device delivers the intermediate activation.
\cref{fig:scheme:sks} illustrates how \sks addresses this problem.
As soon as the backward pass of each block is complete, the parameter updates are performed without waiting for the other devices.
Then, the teacher execution of the next step can start earlier, increasing the training throughput. 
This does not harm the training accuracy by any means because the student blocks have no dependency on the weight parameters of the other blocks, which is a special characteristic of blockwise knowledge distillation training.

\Sks successfully hides the teacher waiting time except for the beginning of each epoch, where full synchronization is needed for validating the whole model.
Because there are usually tens to hundreds of steps per epoch, such overhead is amortized to a negligible amount.

\subsection{\DR}
\label{sec:scheme:dr}
With the \tr and \sks, 
the system throughput is determined by the throughput of the slowest device.
Because of this, load balancing between devices plays a critical role in performance. One straightforward and intuitive load-balancing method is distributing the workload in contiguous blocks.
The distribution is simple because there are only $_{B-1}C_{N-1}$ choices for $B$ blocks and $N$ devices.
Unfortunately, the naive distribution scheme often fails to provide a good balance. 
In \bd, the number of blocks $B$ is determined by the neural network architecture. 
Usually, $B$ is around ten~\cite{resnet, proxyless} and $N$ is four to eight within a single server. 
Because there are not enough number of blocks to distribute to the devices, the naive distribution is likely to end up in a severe workload imbalance.

With \dr, we provide another degree of freedom for workload distribution 
as presented in \cref{fig:scheme:dr}.
Instead of relying on the block granularity, we allow further splitting of each block along the batch dimension.
Thus, when a block is too long, it can be split into two or more smaller effective blocks.
Because a batch is split, the total workload can become larger because of GPU under-utilization.
However, sometimes a slight increase in the total workload is dwarfed by the gain from workload balancing.

\Dr introduces a larger design space to workload distribution, which is difficult to tune manually.
To estimate throughputs of possible schedules,
we measure consumed time of a few test execution for each block under feasible batch sizes. 
Then, considering the practical problem size of both $B$ and $N$ at around ten, the optimal solution can be found using an exhaustive search.
Because the decision is made only once at the beginning, its overhead is amortized over the entire training and is negligible in our experiments.

\section{\scheme Framework}

\subsection{Overall Procedure} 
\begin{algorithm}
\caption{\scheme procedure}\label{alg:procedure}
\begin{algorithmic}[1]
\small
 \renewcommand{\algorithmicrequire}{\textbf{Input:}}
 \Require $ $ \\
 \hspace*{\algorithmicindent}$G$: \# of devices, $D_i$: $i$-th device\\
 \hspace*{\algorithmicindent}$T_i$: Teacher blocks assigned to $D_i$ \\
 \hspace*{\algorithmicindent}$S_i$: Student blocks assigned to $D_i$
  \vspace{2mm}
 \\ \textit{\textbf{Initialization}: Decide $T_i$ and $S_i$ of each device} {\color{blue(ncs)}{// AHD}}
  \vspace{2mm}
\For {each epoch}
  \For {\textbf{parallel} $i = 0, 1,$ ... $,G-1$} 
    \For {each step}
    \IfThen{$D_i$.prev == $\emptyset$}{$act_{in_i}$ = load\_data()}
    \State{\textbf{else} $act_{in_i}$ = receive(from=$D_{i}$.prev) {\color{blue(ncs)}{// TR}}}
    \State $act_{t\_out_i}$ = $T_i$.forward($act_{in_i}$)
    \IfThen{$D_i$.next != $\emptyset$}{send($act_{t\_out_i}$, to=$D_{i}$.next) {\color{blue(ncs)}{// TR}}}
    \State $act_{s\_out_i}$ = $S_i$.forward($act_{in_i}$) 
    \State $S_i$.backward($\mathcal{L}(act_{s\_out_i},act_{t\_out_i}$))  
    \IfThen{AHD\_enabled}{$S_i$.share\_gradient() {\color{blue(ncs)}{// AHD}}} 
    \IfThen{$\sim$DPU\_enabled}{wait\_all\_devices() {\color{blue(ncs)}{// DPU}}}
    \State $S_i$.update\_weight()
  \EndFor
  \EndFor
  \EndFor
 \end{algorithmic} 
 \end{algorithm}

\cref{alg:procedure} displays the overall procedure for \scheme.
At initialization, the optimal schedule is decided from the profiled results, and the blocks are assigned to the devices (line 4).
At the beginning of each step, each device receives the intermediate activation from the previous device (line 9). 
If $T_i$ and $S_i$ contain the first block, the device instead starts with loading the data (line 8). 
It mostly overlaps with the computation except for the first step in each epoch.
After the teacher forward pass is completed (line 10), the result is sent out such that the next device can execute $T_{i+1}$. (line 11).
Then $S_i$ is executed (lines 12-13).
If \dr made a decision to share the block with other devices, gradient sharing is performed (line 14).
Finally, \sks (line 15) removes barrier operation, enabling each device to update its student weight without waiting for the other devices.



\subsection{Implementation}

We used a native PyTorch \texttt{distributed} package for point-to-point communications. 
All communications are implemented to overlap with computations as much as possible.
We used Pytorch \texttt{DistributedDataParallel} class for data-parallel communications.
For \dr, the profiling function is called before training, which runs 100 steps of each block with feasible batch sizes to obtain execution times under the current environment.
Based on these profiled execution times, \scheme determines the best scheduling and starts training.
The implementation of \scheme is available at \url{https://github.com/hongsunjang/Pipe-BD}.

\section{Experimental Setup}
\subsection{Workload}
To demonstrate the advantage of \scheme, we applied it to two popular \bd applications.

\textbf{Neural Architecture Search.}
NAS is the current de facto standard for building a new neural network architecture.
To search for a final architecture, multiple candidate operations
in each layer are associated with a trainable \emph{architecture parameter}, representing the probability of selecting the operation every step.
After the entire network is trained, the operation with the highest probability within each layer is selected as the final architecture.
For an efficient search, \bd is a popular method~\cite{dna,donna} for a smaller solution space.
One notable aspect of NAS is that each step periodically requires two rounds of forward/backward passes for students: one for the architecture parameters and another for the weight parameters.
However, this does not cause any difference to \bd or \scheme because each round can be regarded as a single training step. 
We used ProxylessNAS~\cite{proxyless} as the search backbone.
For the teacher model, we used pre-trained MobileNetV2~\cite{mobilenetv2}.
For other settings, we followed the values suggested from the official implementations of DNA~\cite{dna}.

\textbf{Model Compression.}
Model compression is also another popular application of blockwise knowledge distillation~\cite{pbd, yu2019recasting}. 
A small student neural network model is trained from a larger pretrained teacher model. 
We follow the tradition and use layers of VGG-16~\cite{vgg} as the teacher model and depth-wise separable convolution (DS-Conv)~\cite{mobilenetv1} layers as replacements. We follow the settings from \cite{pbd} for the training.

\subsection{Experimental Environment}
\label{sec:setup}
For the experiments, we use two types of environments.
By default, four RTX A6000 GPUs (Ampere) are attached to an AMD EPYC 7302 CPU.
For additional experiments on a slightly low-cost configuration, four RTX 2080Ti GPUs (Turing) are attached to two Intel Xeon Silver 4214 CPUs.
We used two datasets, CIFAR-10~\cite{cifar} and ImageNet~\cite{imagenet}.
For the model compression, we used stochastic gradient descent (SGD) optimizer with a learning rate of 0.1 for compressing and 0.0001 for finetuning.  
For the NAS, we used SGD optimizer with a learning rate of 0.005 for neural network architecture searching and 0.1 or 0.05 for retraining the final architecture.  

\begin{table}[]
\scriptsize
\centering
\caption{Experimental Environment}\label{tab:environment}
\begin{tabular}{cccc}
\toprule
\multirowcell{8}[-0.4ex]{\textbf {HW}} 
& \multirowcell{4}{Default\\(w/ A6000)} 
& GPU & 4$\times$ NVIDIA RTX A6000 \\
&& CPU & 1$\times$ EPYC 7302, 16 cores \\
&& Memory & 256~GB DDR4 ECC \\
&& Interconnect & PCIe 4.0 \\

\cmidrule(lr){3-4}

& \multirowcell{4}{Alternative\\(w/ 2080Ti)} 
& GPU & 4$\times$ NVIDIA RTX 2080Ti \\
&& CPU & 2$\times$ Xeon 4214 Silver, 12 cores \\
&& Memory & 256~GB DDR4 ECC \\
&& Interconnect & PCIe 3.0 \\

\midrule

\multirowcell{9}[-0.8ex]{\textbf {SW}} 
& \multirow{3}{*}{Common} 
& Python & 3.10\\
&& CUDA & 11.6 \\
&& PyTorch  & 1.13 \\

\cmidrule(lr){3-4}
&\multirow{3}{*}{\makecell{NAS}} 
& Teacher Model & MobileNetV2 \\
&& Kernel Size & 3,5,7 \\
&& Expansion Ratio & 3,6 \\
\cmidrule(lr){3-4}
&\multirow{2}{*}{\makecell{Model\\Compression}} &
 Teacher & VGG-16 \\
&& Replacement & DS-Conv \\

 \bottomrule
\end{tabular}
\end{table}

\subsection{Baselines}
Based on the prior work mentioned in~\cref{sec:related_parallel}, we used two baselines for our experiments.
The first baseline (DP) is the traditional \emph{data-parallel} \bd used in ~\cite{dna} official implementation.
The second baseline (LS) is the \emph{layerwise scheduling} introduced in \cite{pbd}.
Each baseline targets either one of neural architecture search or model compression, so we implemented these baselines to both of our target workloads in PyTorch.

\section{Experimental Results}

\subsection{Speedup and Ablation}

\begin{figure}[t]
\centering
 \includegraphics[width=\columnwidth]{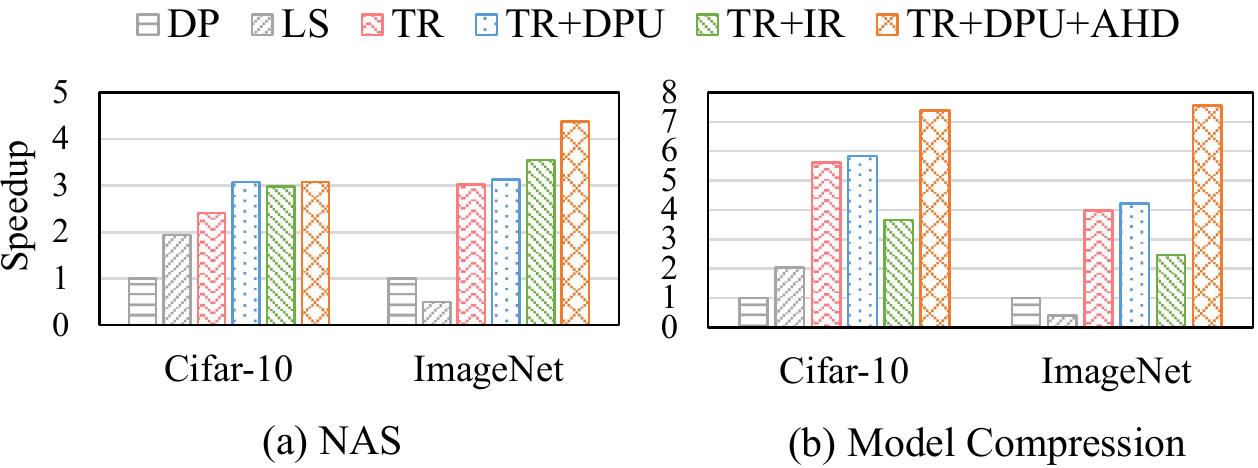}
 \caption{Speedup and ablation of baselines and \scheme. 
 \vspace{-5mm}
 }
 \label{fig:speedup}
\end{figure}

\cref{fig:speedup} shows the speedup of \scheme over the baselines with an ablation study of the proposed techniques using four RTX A6000 GPUs.
Each colored bar shows the speedup of \scheme where 1) only \tr is applied (\textbf{TR}), 2) \sks is further applied (\textbf{TR+DPU}), and 3) all three schemes are applied, including \dr (\textbf{TR+DPU+AHD}).
In addition, we tested an alternative method named \emph{Internal Relaying} (\textbf{TR+IR}). 
With internal relaying, each device trains all existing blocks in every step, and parallelization is obtained via data parallelism.
Instead of re-executing the teacher blocks or relaying activations between devices, the teacher activations are internally stored in memory and are retrieved for the next block.
This approach allows for removing the redundancies of teacher and data loading as well as the load imbalance.
However, it has the disadvantage of using a small batch size per device.
In fact, internal relaying is a special case of \scheme with TR+DPU+AHD when all blocks are only split along the batch dimension.

Among the baselines, LS performs better than DP on Cifar-10 but worse on ImageNet.
Because the composition of the neural networks for ImageNet typically has a few heavy blocks, LS suffers from severe load imbalance.
Nevertheless, they both perform inferior to \scheme. TR provides speedup for all cases due to eliminating extra data loading, redundant teacher execution, and enhancing resource utilization. 
Further, DPU provides additional speedup by removing synchronization barriers, which improves the overlapping of the teacher waiting time with student executions. 
Additionally, AHD removes the pipeline bubbles by balancing workloads, which drives an additional speedup over TR+DPU.

With the ImageNet dataset, its larger spatial dimension of the images (224$\times$224 vs 32$\times$32) leads to heavy workloads in the first block. 
As a result, with TR only, the execution time of block 0 dominates all the others. 
Because of this, DPU has little room for improvement, whereas splitting the workload of the first block with AHD has a large impact on reducing the bubbles.
In contrast, in the Cifar-10 case, the workload is already well-balanced only with TR+DPU version, and the gain from more balancing is offset by the loss from lower resource utilization caused by AHD.

\begin{table*}[t]
\centering
\footnotesize
\caption{Parallel \BD Training Results}\label{tab:accuracy:nas}
\resizebox{\textwidth}{!}
{
\begin{tabular}{lcccccccccccc}
\toprule
\multirowcell{2}[-0.4ex]{\textbf{Task}} & \multirowcell{2}[-0.4ex]{\textbf{Dataset}} & \multicolumn{4}{c}{\textbf{Teacher}} & \multicolumn{4}{c}{\textbf{Student}}& \multicolumn{3}{c}{\textbf{Elapsed Time (1 epoch)}} \\
\cmidrule(lr){3-6}
\cmidrule(lr){7-10}
\cmidrule(lr){11-13}
&& Model & \#Params & FLOPs & Acc. (\%) & Backbone & \#Params & FLOPs & Acc. (\%) & DP & LS & \textbf{\scheme} \\
\midrule
\multirowcell{2}{NAS} & Cifar-10 & MobileNetV2 & 2.24 M & 87.98 M& 95.42\% & ProxylessNAS~\cite{proxyless} & 1.40 M & 76.10 M& 95.48\% & 31.52s. & 16.33s. & \textbf{10.23s.} \\
& ImageNet & MobileNetV2 &3.50 M & 300.77 M& 72.00\% & ProxylessNAS & 4.22 M & 420.20 M & 74.54\% & 62m 21s. & 125m 26s. & \textbf{14m 15s.} \\
\midrule
\multirowcell{2}{Compression} & Cifar-10 & VGG-16 &14.72 M&0.63 B&91.85\%& DS-Conv~\cite{mobilenetv1}&7.25 M &0.39 B&91.51\%& 13m 18s. & 6m 37s. & \textbf{1m 49s.} \\
& ImageNet & VGG-16 &138.36 M&30.98 B&71.59\%& DS-Conv &138.09 M&26.15 B& 71.32\% & 229m 23s. & 566m 49s. & \textbf{60m 39s.} \\
\bottomrule
\end{tabular}
}
 \vspace{-3mm}
\end{table*}

\subsection{Sensitivity and Scheduling}

\begin{figure}[t]
\centering
\begin{subfigure}[t]{\columnwidth}
\centering
\setlength{\abovecaptionskip}{-1mm}
\includegraphics[trim={0mm 0mm 0mm 0mm},clip, width=.9\columnwidth]{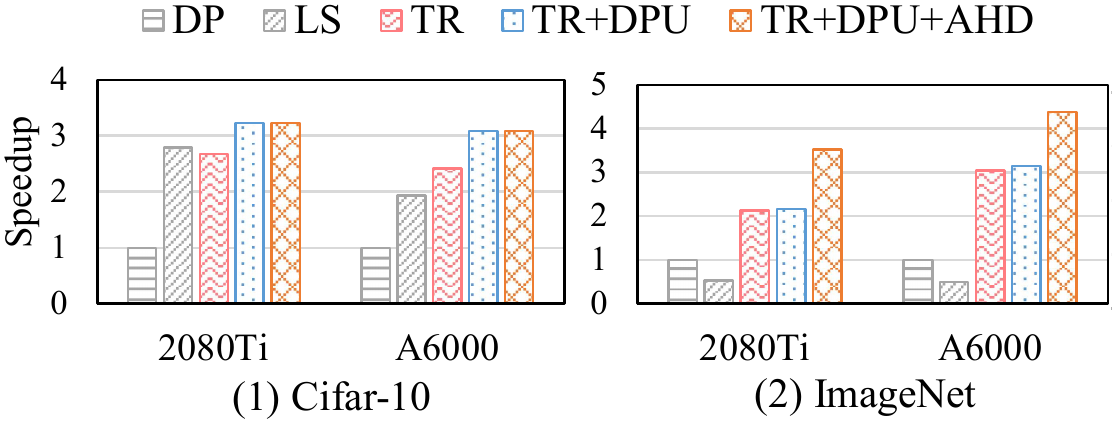}
 \caption{Speedup.}
 \label{fig:gpusensi:spdup}
\end{subfigure}

\begin{subfigure}[t]{0.49\columnwidth}
\centering
\includegraphics[height=1.3cm]{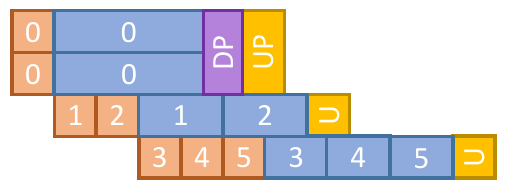}
 \caption{2080Ti schedule.\vspace{-1mm}}
 \label{fig:gpusensi:2080ti}
\end{subfigure}
\begin{subfigure}[t]{0.49\columnwidth}
\centering
\includegraphics[height=1.3cm]{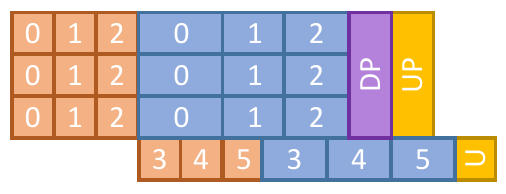}
 \caption{A6000 schedule.\vspace{-1mm}}
 \label{fig:gpusensi:a6000}
\end{subfigure}
\caption{GPU type sensitivity of \scheme on NAS.}
 \vspace{-4mm}
\label{fig:gpusensi}
\end{figure}

\cref{fig:gpusensi}b and \cref{fig:gpusensi}c show how \scheme automatically determines the appropriate schedule according to two different environmental settings with the same NAS on ImageNet workload.
While the speedup trends are similar in \cref{fig:gpusensi}a, they are from different schedules.
The execution time of block 0 is the longest among the six blocks in both settings.
However, the gap is more extensive on A6000 than on 2080Ti.
To mitigate the imbalance, \scheme settles at a schedule where the first three blocks (0-2) are shared on three devices (0-2) for A6000, while block 0 on 2080Ti is shared among two devices (0-1) and two blocks (1-2) are assigned to device 2.

\begin{figure}[t]
\centering
 \includegraphics[width=\columnwidth]{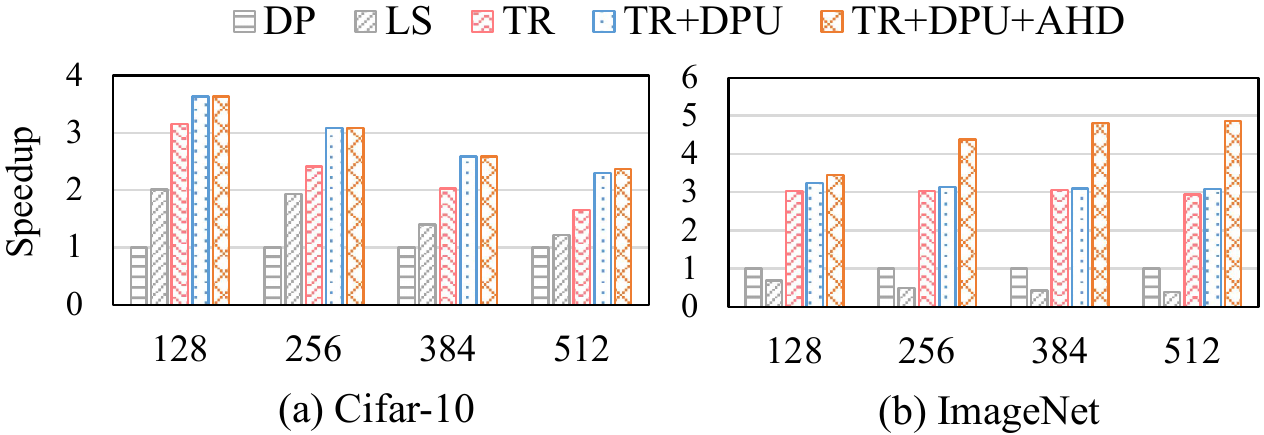}
 \caption{Batch size sensitivity of \scheme on NAS.}
 \vspace{-2mm}
 \label{fig:batchsensi}
\end{figure}

In \cref{fig:batchsensi}, we demonstrate the sensitivity to the batch size on the NAS workload, normalized against DP of each batch size. 
In general, the advantage of \scheme is not very sensitive to the batch size.
One common trend is that the speedup is better in smaller batch sizes because the resource utilization difference becomes more significant with smaller batch sizes.
 One exception is AHD for ImageNet, where the speedup is better on larger batch sizes.
The reason is found in the schedule depicted in \cref{fig:gpusensi:a6000} which uses three-way data parallelism to balance workloads.
Because the training time for the student is shorter in both the baseline and \scheme with larger batch sizes, reduction in the teacher redundancy and extra data loading account more for the overall speedup.

\subsection{Memory Overhead}
\begin{figure}[t]
\centering
 \includegraphics[width=\columnwidth]{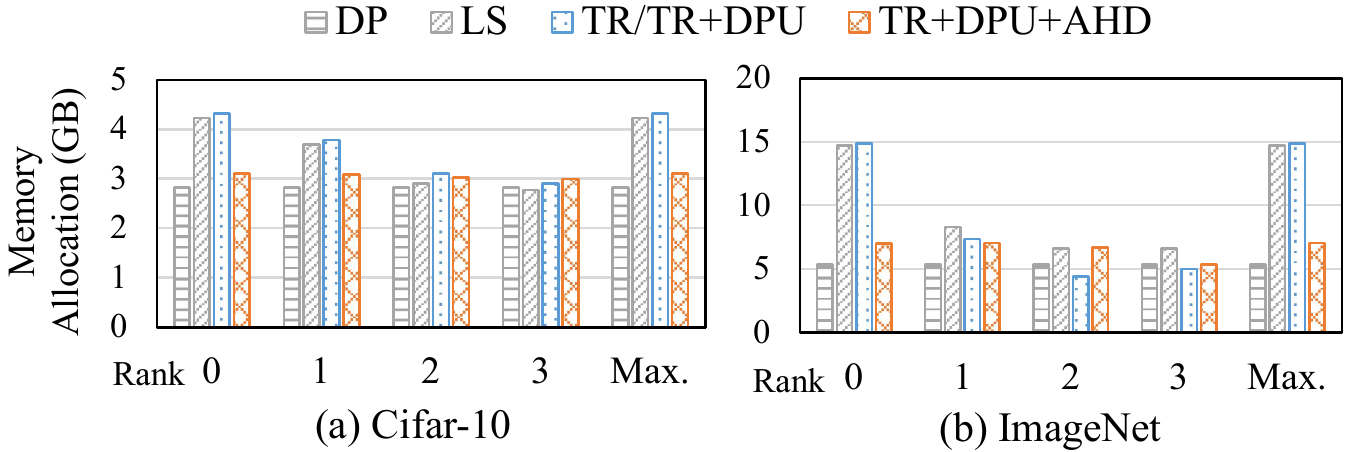}
 \caption{Memory overhead of \scheme on NAS.}
 \vspace{-2mm}
 \label{fig:memory}
\end{figure}

\cref{fig:memory} depicts the memory overhead of \scheme on the NAS task for each rank (GPU).
Due to the characteristics of CNN-based models, lower-indexed teacher blocks generally have larger feature map sizes.
TR and DPU consume more memory than DP because of this characteristic, especially on rank 0.
This outcome is also demonstrated in~\cref{fig:memory}b because models for ImageNet contain even larger feature map sizes in lower-indexed blocks. 
However, AHD successfully addresses this issue using data parallelism in a hybrid manner, which lessens the memory overhead of earlier ranks as depicted in~\cref{fig:gpusensi:a6000}.
As a result, \scheme provides superior multi-fold speedups with a minor 8.7\% and 21.3\% additional memory overheads over DP on average for Cifar-10 and ImageNet, respectively.

\subsection{Training Quality}

\scheme has no component that can hurt the accuracy because it only alters the scheduling strategy. 
Nonetheless, we report the accuracy in \cref{tab:accuracy:nas} to demonstrate that the \scheme framework faithfully reproduces the end-to-end training results in the prior art, with much shorter training time.
For all use cases under evaluation, \scheme achieves significant speedup with the same accuracy. 
\JL{is there any better way to put this?}
\HS{ Accuracy is not harmed by Pipe-BD.. explain!}

\section{Conclusion} 
We propose \scheme, a novel parallelization method for \bd.
By restructuring the existing parallelization scheme, we achieve a multi-fold speedup on various use cases.
In this study, we focused on a single-node, multi-GPU setting since it is the most common setup. 
However, if the method has to be scaled for a multi-node setting, the communication overhead needs to be addressed.
Along with the heterogeneous GPU/servers, this will be our future direction.

\footnotesize
\section*{Acknowledgement}
\vspace{3mm}

This work was partly supported by
the National Research Foundation of Korea (NRF) grants (2022R1C1C1011307, 2022R1C1C1008131) and Samsung Electronics Co., Ltd (IO221213-04119-01) and
Institute of Information \& communications Technology Planning \& Evaluation (IITP) grants (2020-0-01361) funded by the Korean government (MSIT).
\vspace{4mm}

\def\bibfont{\scriptsize}
\bibliographystyle{IEEEtran}

\bibliography{refs}

\end{document}

%% file: packages.tex
\usepackage{amsmath,amssymb,amsfonts}
\usepackage{nccmath} 
\usepackage[numbers]{natbib}
\usepackage[utf8]{inputenc}

\usepackage{algorithm}
\usepackage{algpseudocode}
\algnewcommand{\IfThen}[2]{
  \State \algorithmicif\ #1\ \algorithmicthen\ #2\ }
\algnewcommand{\IfThenElse}[3]{
  \State \algorithmicif\ #1\ \algorithmicthen\ #2\ \\ \algorithmicelse\ #3\ }

\usepackage{textcomp}

\usepackage[hyphens]{url}

\usepackage[bookmarks=true,breaklinks=true,letterpaper=true,colorlinks,linkcolor=black,citecolor=blue,urlcolor=black]{hyperref}
\usepackage[capitalize]{cleveref} 
\usepackage{subcaption}
\usepackage{graphicx}
\usepackage{verbatim}
\usepackage{nicefrac}
\usepackage{siunitx}
\usepackage{float} 

\usepackage[export]{adjustbox} 

\usepackage{balance}

\usepackage{soul, color}

\usepackage{xcolor} 
\usepackage{booktabs}
\usepackage{multicol}
\usepackage{multirow}
\usepackage{makecell}
\usepackage{xspace}
\usepackage{marginnote}
\usepackage{setspace} 

\usepackage{pgfplots}
\usepackage{pgfplotstable}
\usepgfplotslibrary{groupplots}
\usepackage{tikz}
\usetikzlibrary{patterns}
\usetikzlibrary{backgrounds}

\usepackage{comment}
\usepackage{marginnote}

\usepackage{listings}

\usepackage{pifont}

\usepackage[linewidth=1pt]{mdframed}
\usepackage{tcolorbox}

\usepackage{enumitem}

%% file: commands.tex
\definecolor{olivegreen}{rgb}{0, 0.6, 0}
\definecolor{grannysmithapple}{rgb}{0.66, 0.89, 0.63}
\definecolor{ceruleanblue}{rgb}{0.16, 0.32, 0.75}
\definecolor{babyblue}{rgb}{0.54, 0.81, 0.94}
	
\newcommand{\JL}[1]{{\color{olivegreen}[\textbf{\sc JLee}: \textit{#1}]}}
\definecolor{blue(ncs)}{rgb}{0.0, 0.53, 0.74}
\definecolor{blush}{rgb}{0.87, 0.36, 0.51}

\definecolor{carminered}{rgb}{1.0, 0.0, 0.22}

\newcommand{\HS}[1]{{\color{orange}[\textbf{\sc HS}:\textit{#1}]}}

\newcommand{\JS}[1]{{\color{blue(ncs)}[\textbf{\sc JS}: \textit{#1}]}}

\def\final{}   
\ifdefined\final
\renewcommand{\JL}[1]{}
\renewcommand{\JS}[1]{}
\renewcommand{\HS}[1]{}

\fi

\newcommand{\scheme}{Pipe-BD\xspace}
\newcommand{\TR}{Teacher Relaying\xspace}

\newcommand{\tr}{teacher relaying\xspace}

\newcommand{\SKS}{Decoupled Parameter Update\xspace}
\newcommand{\Sks}{Decoupled parameter update\xspace}
\newcommand{\sks}{decoupled parameter update\xspace}

\newcommand{\DR}{Automatic Hybrid Distribution\xspace}
\newcommand{\Dr}{Automatic hybrid distribution\xspace}
\newcommand{\dr}{automatic hybrid distribution\xspace}

\newcommand{\BD}{Blockwise Distillation\xspace}
\newcommand{\Bd}{Blockwise distillation\xspace}
\newcommand{\bd}{blockwise distillation\xspace}

\usepackage{soul}

\definecolor{black}{HTML}{000000}
\definecolor{white}{HTML}{ffffff}
\definecolor{color1}{HTML}{90ee90}
\definecolor{color2}{HTML}{F0E68C}
\definecolor{color3}{HTML}{DCD0FF}
\definecolor{color4}{HTML}{B19CD9}
\definecolor{color5}{HTML}{FFB6C1}
\definecolor{color6}{HTML}{20B2AA}
\definecolor{color7}{HTML}{87CEEB}
\definecolor{color8}{HTML}{FFA07A}

\pgfplotsset{compat=newest}
\makeatletter
\newcommand\resetstackedplots[1]{
\makeatletter
\pgfplots@stacked@isfirstplottrue
\makeatother
\pgfplotstablenew[
  create on use/x/.style={create col/expr={\pgfplotstablerow}},
  create on use/y/.style={create col/expr={0}},
  columns={x,y}]{#1}\zerotable
\addplot [forget plot,draw=none] table[x=x,y=y] from \zerotable;
}
\makeatother